\documentclass[a4paper,twoside]{article}

\usepackage{epsfig}
\usepackage{subcaption}
\usepackage{calc}
\usepackage{amssymb}
\usepackage{amstext}
\usepackage{amsmath}
\usepackage{amsthm}
\usepackage{multicol}
\usepackage{pslatex}
\usepackage{apalike}
\usepackage{algorithm2e}
\usepackage{array}   
\usepackage{graphicx}
\usepackage{latexsym}
\usepackage{float}
\usepackage{multirow}
\usepackage{colortbl}
\usepackage{xcolor}
\usepackage{dblfloatfix}
\usepackage{tabularx}

\usepackage{mathtools}
\usepackage{amsfonts}
\usepackage[bottom]{footmisc}
\usepackage{hyperref}
\usepackage{SCITEPRESS}     

\newcommand{\norm}[1]{\lVert #1 \rVert}
\newcommand{\abs}[1]{\lvert #1 \rvert}
\newcolumntype{C}{>{$}c<{$}} 

\begin{document}

\title{Investigating the Corruption Robustness of Image Classifiers with Random $p$-norm Corruptions}

\author{\authorname{Georg Siedel\sup{1,2}, Weijia Shao\sup{1}, Silvia Vock\sup{1}, Andrey Morozov\sup{2}}
\affiliation{\sup{1}Federal Institute for Occupational Safety and Health (BAuA), Dresden, Germany}
\affiliation{\sup{2}University of Stuttgart, Germany}
\email{\{siedel.georg, shao.weijia, vock.silvia\}@baua.bund.de, andrey.morozov@ias.uni-stuttgart.de}
}

\keywords{image classification, corruption robustness, $p$-norm}

\abstract{Robustness is a fundamental property of machine learning classifiers required to achieve safety and reliability. In the field of adversarial robustness of image classifiers, robustness is commonly defined as the stability of a model to all input changes within a $p$-norm distance. However, in the field of random corruption robustness, variations observed in the real world are used, while $p$-norm corruptions are rarely considered. This study investigates the use of random $p$-norm corruptions to augment the training and test data of image classifiers. We evaluate the model robustness against imperceptible random $p$-norm corruptions and propose a novel robustness metric. We empirically investigate whether robustness transfers across different $p$-norms and derive conclusions on which $p$-norm corruptions a model should be trained and evaluated. We find that training data augmentation with a combination of $p$-norm corruptions significantly improves corruption robustness, even on top of state-of-the-art data augmentation schemes.}

\onecolumn \maketitle \normalsize \setcounter{footnote}{0} \vfill

\section{\uppercase{Introduction}}
State-of-the-art computer vision models achieve human-level performance in various tasks, such as image classification \cite{Krizhevsky2017}. This makes them potential candidates for challenging vision tasks. However, they tend to be easily fooled by small changes in their input data, which limits their overall dependability to perform in safety-critical applications \cite{Carlini2016}. For classification models, robustness against small data changes is therefore considered a fundamental pillar of AI dependability and has attracted considerable research interest in recent years. 

Within the robustness research landscape, the adversarial robustness domain has received the most attention \cite{Drenkow2021}. An adversarial attack aims to find worst-case counterexamples for robustness. However, vision models are not only vulnerable to small worst-case data manipulations in the input data but also to randomly corrupted input data \cite{Dodge2017}. Accordingly, the corruption robustness\footnote{Also called statistical robustness} domain aims at models that perform similarly well on data corrupted with random noise. 
A clear distinction needs to be made between adversarial robustness and corruption robustness: Existing research suggests that these two types of robustness target different model properties and applications \cite{Wang2021}. Also, adversarial robustness and corruption robustness do not necessarily transfer to each other, and adversarial robustness is much harder to achieve in high-dimensional input space \cite{Fawzi2018a,Ford2019}.

\begin{figure}[t]
\centerline{\includegraphics[width=75mm, trim={2mm 3mm 1mm 2mm},clip]{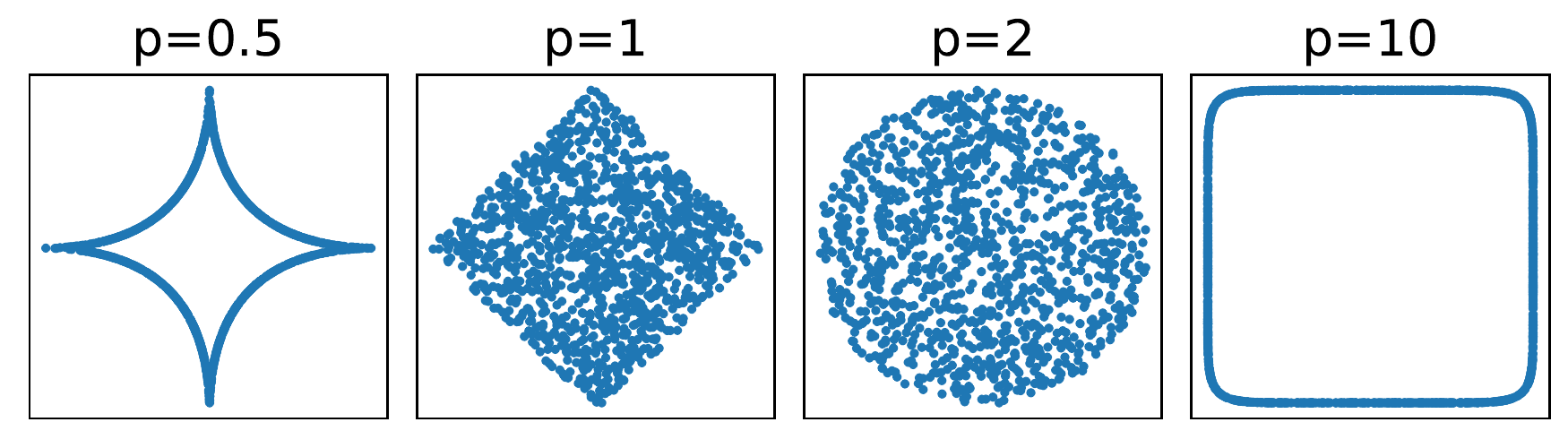}}
\caption{Samples drawn uniformly in 2D from a $L_{0.5}$ and a $L_{10}$ norm sphere (left and right) and a $L_1$ and a $L_2$ norm ball (middle).} \label{lp-2d}
\vspace{3pt}
\end{figure}

\subsection{Motivation}
The adversarial robustness domain provides the clearest definition of robustness based on a maximum manipulation distance in the $p$-norm space (see Section 2.1) \cite{Drenkow2021}. Accordingly, there exists a set of $p$-norms and maximum distances $\epsilon$ for each norm that are commonly used for robustness evaluation on popular image classification benchmarks, e.g. $L_\infty$ with $\epsilon=8/255$ and $L_2$ with $\epsilon=0.5$ on the CIFAR and SVHN benchmark datasets \cite{Croce2020,Yang2020}.

In contrast, corruption robustness is commonly assessed by testing data corruptions originating from camera, hardware, or environment \cite{Hendrycks2019a}. Such data corruptions can occur in real-world applications\footnote{thus we refer to them as "real-world corruptions", even though they are usually artificially created}. Investigating robustness using $p$-norm distances is rarely employed in the corruption robustness domain. We present three arguments to motivate our investigation into this area. 

First of all, the adversarial robustness targets the significant performance degradation caused by $p$-norm manipulations that are small or, even worse, imperceptible \cite{Carlini2016,Szegedy2013,Zhang2019}. From our point of view, a classifier should generalize in line with human perception, regardless of whether the manipulations are adversarial or random. Therefore, we want to investigate whether the same is true for random $p$-norm corruptions.

Second, related work has shown that increasing the range of training data augmentations can improve the performance of classifiers \cite{Mintun2021,Muller2021}. Therefore, we expect that combining different $p$-norm corruptions at training time will lead to effective robustness improvements.

Furthermore, studies have shown the difficulty of predicting transferability among different types of corruption \cite{Hendrycks2019a,Ford2019}. Given the large differences between the volumes covered by different $p$-norm balls in high-dimensional space (see Table~\ref{volumefactors} in the appendix), we suspect the same for different $p$-norm corruptions, which should be observable in the empirical evaluation. 

\subsection{Contributions}
This paper investigates image classifiers trained and tested on random $p$-norm corruptions.\footnote{Code available: 
\href{https://github.com/Georgsiedel/Lp-norm-corruption-robustness}{https://github.com/Georgsiedel/Lp-norm-corruption-robustness}}
The main contributions can be summarized as follows:

\begin{itemize}

\item We test classifiers on random quasi-imperceptible corruptions from different $p$-norms and demonstrate a performance degradation. We propose to measure this minimum requirement for corruption robustness with a corresponding robustness metric.

\item We present an empirically effective training data augmentation strategy based on combinations of $p$-norm corruptions that improve robustness more effectively than individual corruptions.

\item We evaluate how robustness obtained from training on $p$-norm corruptions transfers to other $p$-norm corruptions and to real-world corruptions. We discuss the transfer of $p$-norm corruption robustness from a test coverage perspective. 
\end{itemize}

\section{\uppercase{Preliminaries}}

\subsection{Robustness definition} \quad A classifier $g$ is locally robust at a data point $x$ within a distance $\epsilon>0$, if $g(x)=g(x')$ holds for all perturbed points $x'$ that satisfy 
\begin{equation}
\operatorname{dist}(x, x')\leq\epsilon
\label{thesystem}
\end{equation}
with $x'$ close to $x$ according to a predefined distance measure \cite{Carlini2016}. 

In the adversarial attack domain \cite{Carlini2016}, this distance in $\mathbb{R}^d$ is commonly induced by a $p$-norm \cite{Weng2018,Yang2020}, for $0\leq p\leq \infty$ \footnote{Here, we abuse the term "norm distance" by also using $0 \leq p<1$, which are no norms by mathematical definition.}:
\begin{equation}
\norm{x-x'}_p=(\sum\nolimits_{i=1}^{d}\abs{x_i-x'_i}^{ p })^{ 1/p }.
\end{equation}

\subsection{Sampling algorithm}
In order to experiment with random corruptions uniformly distributed within a $p$-norm ball or sphere, an algorithm that scales to high-dimensional space is required for all norms $0 < p < \infty$\footnote{$p=0$ (set a ratio $\epsilon$ of dimensions to 0 or 1) and $p=\infty$ (add uniform random value from [$\epsilon$, $-\epsilon$] to every dimension) are trivial from a sampling perspective.}. We use the approach in \cite{Calafiore1998}, which is described in detail in the Appendix.

Figure \ref{lp-2d} visualises samples for different $p$-norm unit balls in $\mathbb{R}^2$ using the proposed algorithm, demonstrating its ability to sample uniformly inside the ball and on the sphere.

\section{\uppercase{Related Work}}

\textbf{$p$-norm Distances in the Robustness Context} \quad As described in Section 2.1, $p$-norm distances are used to define adversarial robustness. Furthermore, $p$-norm distances have been used to measure certain properties of image data related to robustness. One such property is the threshold at which an image manipulation is imperceptible, which is a reference point in the field of adversarial robustness \cite{Szegedy2013}. $L_\infty$ manipulations with $\epsilon=8/255$ have been used in conjunction with the imperceptibility threshold \cite{Zhang2019,Madry2017}. However, it is difficult to justify an exact imperceptibility threshold \cite{Zhang2019}. Researchers have therefore called for further research into "distance metrics closer to human perception" \cite{Huang2020}. There exist similarity measures for images that are more aligned with human perception than $p$-norm distances \cite{Wang2004}. To the best of our knowledge, these measures have not yet been used as distance measures for the robustness evaluation of perception models.

Another property of image data measured using $p$-norm distances in the context of robustness is the class separation of a dataset. The average class separation is used by \cite{Fawzi2018b} in order to obtain a comparable baseline distance for robustness testing. Similarly, \cite{Yang2020} measure the minimum class separation on typical image classification benchmarks using $L_\infty$-distances. The authors conclude the existence of a perfectly robust classifier with respect to this distance. Building on this idea, \cite{Siedel2022} investigate robustness on random $L_\infty$-corruptions with minimum class separation distance, obtaining an interpretable robustness metric.

\cite{Wang2021} test and train image classifiers on data corrupted by few random $L_\infty$ corruptions. Overall, little research has been directed at using random corruptions based on $p$-norm distances to evaluate robustness.

\textbf{Real-world Data Corruptions for Testing} \quad In comparison, much research uses real-world corruptions to evaluate robustness, such as the popular benchmark by \cite{Hendrycks2019a} and extensions thereof like in \cite{Mintun2021}.

\textbf{Real-world Data Corruptions for Training} \quad Real-world data corruptions are also used as training data augmentations alongside geometric transformations in order to obtain better generalizing or more robust classifiers \cite{Shorten2019}. Some training data augmentations only target accuracy but not robustness \cite{Cubuk2020,Muller2021}. A few approaches use only geometric transformations, such as cuts, translations and rotations, to improve overall robustness \cite{Yun2019,Hendrycks2019b}. Some approaches make use of random noise, such as Gaussian and Impulse noise, to improve robustness \cite{Lopes2019,Dai2021,Lim2021,Erichson2022}.

Even though accuracy and robustness have long been considered as an inherent trade-off \cite{Tsipras2019,Zhang2019}. Several described data augmentation methods manage to increase corruption robustness along with accuracy \cite{Hendrycks2019b,Lopes2019}. Some data augmentation methods even leverage random corruptions to implicitly or explicitly improve or give guarantees for adversarial robustness \cite{Cohen2019,Weng2019,Lecuyer2019,Yun2019}.

\textbf{Classic Noise and $p$-norm Corruptions} \quad Some perturbation techniques used in signal processing are closely related to $p$-norm corruptions. Gaussian noise shares similarities with $L_2$-norm corruptions \cite{Cohen2019}. Impulse Noise or Salt-and-Pepper Noise are similar to this paper's notion of $L_0$-norm corruptions. Applying brightness or darkness to an image is a non-random $L_\infty$ corruption that applies the same change to every pixel. This similarity further motivates our investigation of the behaviour of various $p$-norm corruptions.

\textbf{Robustness Transferability} \quad Robustness is typically specific to the type of corruption or attack the model was trained for. Adversarial or corruption robustness is only to a limited extent transferable between each other \cite{Fawzi2018a,Fawzi2018b,Rusak2020}, across different $p$-norm attacks and attack strengths \cite{Carlini2019}, or across real-world corruption types \cite{Hendrycks2019a,Ford2019}. This finding motivates our investigation into whether random $p$-norm corruption robustness transfers to other $p$-norm corruptions and real-world corruptions.

\textbf{Wide Training Data Augmentation} \quad Recent efforts suggest that choosing randomly from a wide range of augmentations can be more effective compared to more sophisticated augmentation strategies \cite{Mintun2021,Muller2021}. Furthermore, \cite{Kireev2022} find that training with single types of noise overfits with regards to both noise type and noise level. These results encourage us to investigate the combination of training time $p$-norm corruptions that effectively improve corruption robustness.

\section{\uppercase{Experimental Setup}}
\subsection{Robustness Metrics}

In addition to the standard test error $E_{clean}$, we use 4 metrics to assess the corruption robustness of all our trained models. For all reported metrics, low values indicate a better performance.

To cover well-known real-world corruptions, we compute the \textbf{$\mathbf{mCE}$ (mean Corruption Error)} metric using the benchmark by \cite{Hendrycks2019a}. We use a 100\% error rate as a baseline, so that $\mathbf{mCE}$ corresponds to the average error rates $\mathbf{E}$ across 19 different corruptions $\mathbf{c}$ and 5 corruption severities $\mathbf{s}$ each:
\begin{equation}
    \mathrm{mCE}=(\sum\nolimits_{s=1}^{5}\sum\nolimits_{c=1}^{19}E_{s,c})/(5*19)
\end{equation}
We additionally report $\mathbf{mCE}$ without the 4 noise corruptions included in the metric, which we denote \textbf{$\mathbf{mCE_{xN}}$} (mean Corruption Error ex Noise). This metric evaluates robustness against the remaining 15 corruption types that are not based on any form of pixel-wise noise, such as the $p$-norm corruptions we train on. 
To cover random $p$-norm corruptions, we introduce a robustness metric \textbf{$\mathbf{mCE_{L_p}}$} (mean Corruption Error $p$-norm), which is calculated similarly as (3) from the average of the error rates. As shown in Table \ref{corr-sets}, the corruptions $\mathbf{c}$ for $\mathbf{mCE_{L_p}}$ are 9 different $p$-norms and the severities $\mathbf{s}$ are 10 different $\epsilon$-values for each $p$-norm. The $p$-norms and $\epsilon$-values were manually selected to cover a wide range of values and to lead to significant and comparable performance degradation of a standard model.

\begin{figure}[t]
\centering
\begin{subfigure}[b]{0.3\textwidth}
    \centerline{\includegraphics[width=75mm, trim={4.6mm 89.5mm 3.7mm 96mm},clip]{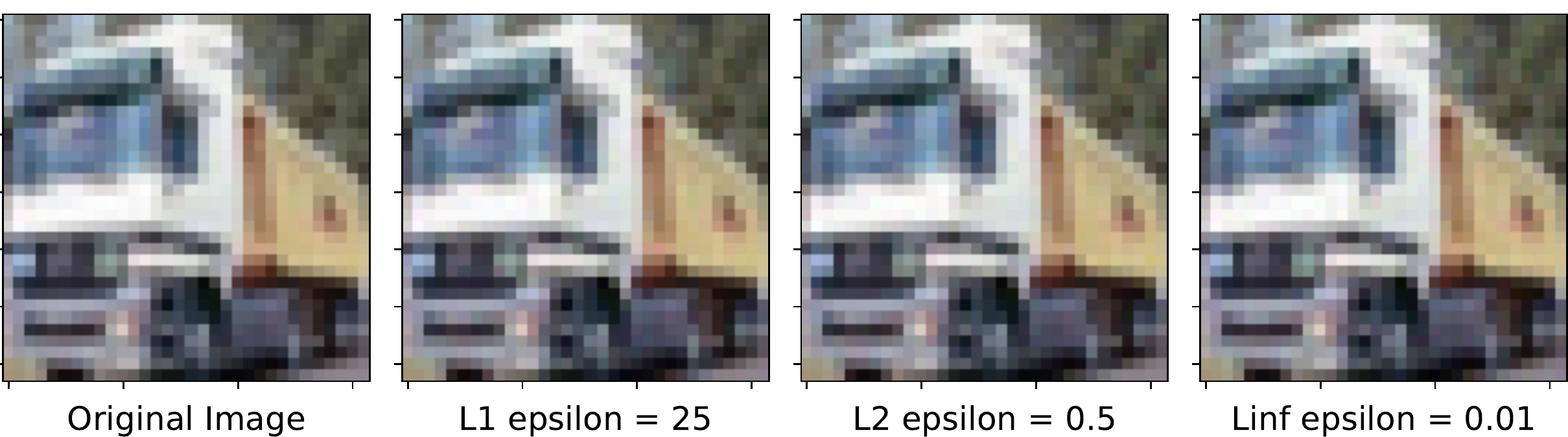}}
   \label{imperceptible_corruptions} 
\end{subfigure}
\begin{subfigure}[b]{0.3\textwidth}
    \centerline{\includegraphics[width=75mm, trim={4.6mm 89.5mm 3.7mm 96mm},clip]{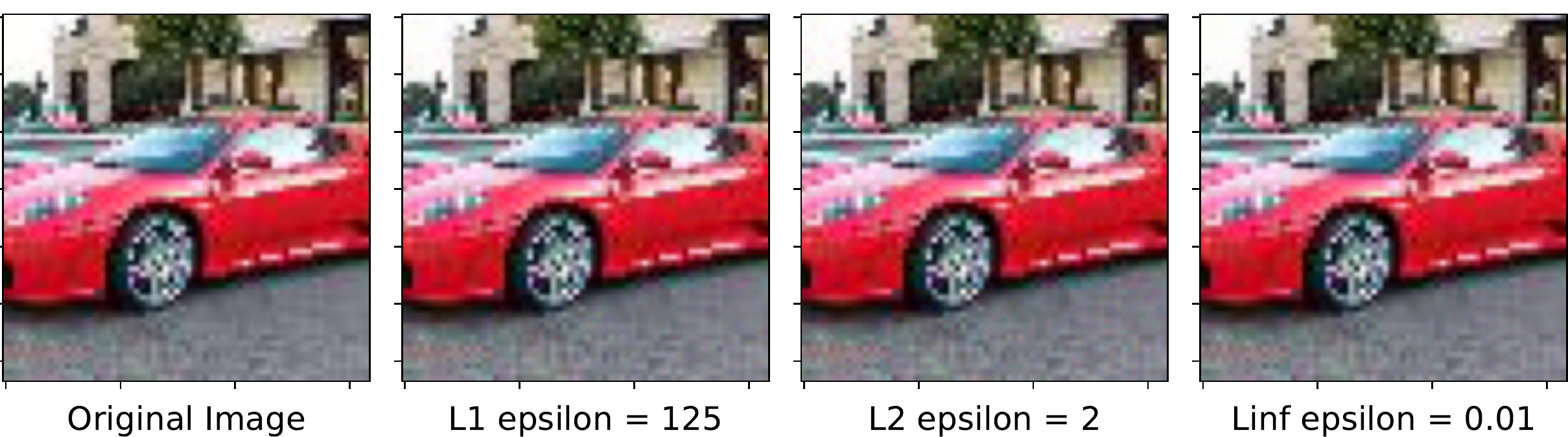}}
   \label{fig:Ng1} 
\end{subfigure}
\caption{Examples from the chosen set of imperceptible corruptions on CIFAR (above) and TinyImageNet (below)} \label{impercept}
\vspace{7pt}
\end{figure}

In order to investigate imperceptible random $p$-norm corruptions, we propose the \textbf{$\mathbf{iCE}$} (imperceptible Corruption Error) metric:
\begin{equation}
\label{eqn:iCE}
    iCE=(\sum\nolimits_{i=1}^{n}E_i-E_{clean})/(n * E_{clean}),
\end{equation}
where $n$ is the number of different imperceptible corruptions. This metric can be considered as a minimum requirement for the corruption robustness of a classifier. $iCE$ quantifies the increase in error rate relative to the clean error rate $E_{clean}$ and is therefore expressed in $\%$. For each dataset in our study, we choose $n=6$ $p$-norm corruptions (Table~\ref{corr-sets}). We select $p$ and $\epsilon$-values based on a small set of randomly sampled and maximally corrupted images. Figure~\ref{impercept} visualizes such corruptions compared to the original image for the CIFAR and TinyImageNet datasets. 

\begin{table}
\begin{center}
\setlength\tabcolsep{4pt}
{\caption{Sets of $p$-norm corruptions for calculating the $mCE_{L_p}$ and $iCE$ metrics. Brackets contain the minimum and maximum out of 10 $\epsilon$-values for each $p$-norm. For $L_0$, $\epsilon$ is the ratio of maximally corrupted image dimensions.}\label{corr-sets}}
\resizebox{\linewidth}{!}{%
\begin{tabular}{lcccc}
\\[-10pt]
\hline
\rule{0pt}{10pt}
$p$&\multicolumn{2}{c}{$mCE_{L_p}$ [$\epsilon_{min}, \epsilon_{max}$]}&\multicolumn{2}{c}{$iCE$ [$\epsilon$]}\\[3pt]
\hline
\rule{0pt}{11pt}
&CIFAR&TIN&CIFAR&TIN\\[2pt]
\cline{2-5}
\\[-8pt]
$0$&[0.005, 0.12]& [0.01, 0.3]& &\\
0.5& [2.5e+4, 4e+5]& [2e+5, 1.2e+7]&2.5e+4&7e+5\\
1& [12.5, 200]& [37.5, 1500]&25&125\\
2& [0.25, 5]& [0.5, 20]&0.5&2\\
5& [0.03, 0.6]& [0.05, 1.5]& & \\
10& [0.02, 0.3]& [0.02, 0.7]&0.03&0.06\\
50& [0.01, 0.18]& [0.02, 0.35]&0.02&0.04\\
200& [0.01, 0.15]& [0.02, 0.3]&&\\
$\infty$& [0.005, 0.15]& [0.01, 0.3]&0.01&0.01\\
\\[-9pt]
\hline
\end{tabular}}
\end{center}
\vspace{2pt}
\end{table}

\subsection{Training Setup}
Experiments are performed on the CIFAR-10 (C10), CIFAR-100 (C100) \cite{krizhevsky2009} and Tiny ImageNet (TIN) \cite{Le2015} classification datasets. We train 3 architectures of convolutional neural networks: A WideResNet28-4 (WRN) with $0.3$ dropout probability \cite{Zagoruyko2016}, a DenseNet201-12 (DN) \cite{Huang2017} and a ResNeXt29-32x4d (RNX) \cite{Xie2017} model. We use a Cosine Annealing Learning Rate schedule with $150$ epochs and an initial learning rate of $0.1$, restarting after $10$, $30$ and $70$ epochs \cite{Loshchilov2016}. As an optimizer, we use SGD with $0.9$ momentum and $0.0005$ learning rate decay. We use a batch size of $384$, resize the image data to the interval [0,1] and augment with random horizontal flips and random cropping. 

In addition to standard training, we apply a set of $p$-norm corruptions to the training dataset, each with different $\epsilon$-values. The norms used are $L_{0}, L_{0.5}, L_1, L_2, L_{50}$ and $L_\infty$. $L_1(40)$ denotes a model trained exclusively on data augmented with $L_1$ corruptions with $\epsilon\leq40$. We manually selected the set of $p$-norms to cover a wide range of values and the 2 $\epsilon$-values for each $p$-norm so that we observe a meaningfully large effect on the metrics for all $p$-norms.

We also train with combinations of $p$-norm corruptions by randomly applying one from a wider set of $p$-norm corruptions and $\epsilon$-values. An overview of the 3 different combined corruption sets is shown in Table \ref{combined-corr}, denoted C1, C2 and C3. We apply a $p$-norm corruption sample to a minibatch of 8 images during training. This avoids computational drawbacks of calculating the sampling algorithm for $0 < p < \infty$ on the CPU, resulting in almost no additional overhead.

\begin{table}
\begin{center}
\setlength\tabcolsep{4.5pt}
{\caption{Sets of $p$-norm corruptions for the training of the models C1, C2 and C3. Brackets contain the minimum and maximum out of 10 (C1 and C2) or 5 (C3) $\epsilon$-values for each $p$-norm.}\label{combined-corr}}
\resizebox{\linewidth}{!}{%
\begin{tabular}{lcccc}
\\[-10pt]
\hline
\rule{0pt}{10pt}
$p$&\multicolumn{4}{c}{[$\epsilon_{min}$, $\epsilon_{max}$]}\\[3pt]
\hline
\rule{0pt}{11pt}
&C1&C2&C3 on CIFAR&C3 on TIN\\[2pt]
\cline{2-5}
\\[-8pt]
$0$&$+$&$+$&[0.005, 0.03]$*$&[0.01, 0.075]$*$\\
0.5&$+$&&[2.5e+4, 1.5e+5$*$&[2e+5, 1.8e+6]$*$\\
1&$+$&&[12.5, 75]$*$&[37.5, 300]$*$\\
2&$+$&$+$&[0.25, 1.5]$*$&[0.5, 4]$*$\\
5&$+$&&[0.03, 0.2]$*$&[0.05, 0.3]$*$\\
10&$+$&&[0.02, 0.1]$*$&[0.02, 0.14]$*$\\
50&$+$&&[0.01, 0.06]$*$&[0.02, 0.1]$*$\\
200&$+$&&[0.01, 0.05]$*$&[0.02, 0.08]$*$\\
$\infty$&$+$&$+$&[0.005, 0.04]$*$&[0.01, 0.06]$*$\\
\\[-9pt]
\hline
\\[-13pt]
\end{tabular}}
\end{center}
\footnotesize{$+$ Same $\epsilon$-values as in $mCE_{L_p}$ for this $p$-norm (see Table \ref{corr-sets}}\\
\footnotesize{$*$ Lowest 5 of the 10 $\epsilon$-values in $mCE_{L_p}$ for this $p$-norm (see Table \ref{corr-sets}}\\
\\[-13pt]
\end{table}

Furthermore, the $p$-norm corruption combinations are compared and combined with the 4 state-of-the-art data augmentation and mixing strategies TrivialAugment (TA) \cite{Muller2021}, RandAugment (RA) \cite{Cubuk2020} and AugMix (AM) \cite{Hendrycks2019b} and Mixup (MU) \cite{Zhang2018}. Accordingly, TA$+$C1 denotes a model trained on images with TA augmentation followed by the C1 corruption.

\section{\uppercase{Results}}

\begin{table}
\begin{center}
\setlength\tabcolsep{3pt}
{\caption{All metrics for the DenseNet model on CIFAR-100. We compare standard training data with various $p$-norm corrupted training data and combinations of $p$-norm corrupted training data.}\label{p-norm-table}}
\resizebox{\linewidth}{!}{%
\begin{tabular}{lccccc}
\\[-10pt]
\hline
\rule{0pt}{11pt}
Model&$E_{clean}$&$mCE$&$mCE_{xN}$&$mCE_{L_p}$&$iCE$\\[3pt]
\hline
\\[-6pt]
Standard&\textbf{23.21}&51.33&46.19&53.70&12.41\%\\[7pt]
$L_{0}(0.01)$&24.08&47.32&45.76&45.57&8.6\%\\
$L_{0}(0.03)$&23.62&47.24&44.80&47.25&7.7\%\\
$L_{0.5}$(7.5e+4)&25.86&47.91&45.83&41.22&\textbf{-1.0\%}\\
$L_{0.5}$(1.5e+5)&26.35&43.11&44.15&33.83&-0.1\%\\
$L_{1}(50)$&25.68&47.78&45.11&41.86&\textbf{-1.0\%}\\
$L_{1}(100)$&29.12&45.64&46.28&34.12&-0.1\%\\
$L_{2}(1)$&24.77&48.92&45.71&44.06&0.0\%\\
$L_{2}(2.5)$&28.98&46.33&46.39&35.11&-0.4\%\\
$L_{50}(0.03)$&24.82&48.67&45.27&44.87&-0.7\%\\
$L_{50}(0.08)$&28.95&47.02&46.68&36.39&-0.5\%\\
$L_\infty(0.02)$&25.05&48.63&45.47&43.79&-0.6\%\\
$L_\infty(0.04)$&28.89&47.69&47.04&37.41&-0.2\%\\[7pt]
C1&27.51&39.66&41.85&29.67&0.8\%\\
C2&26.04&\textbf{39.31}&41.75&\textbf{28.38}&0.9\%\\
C3&24.60&40.30&\textbf{41.73}&29.88&0.7\%\\
\\[-9pt]
\hline
\end{tabular}}
\end{center}
\vspace{3pt}
\end{table}

Table \ref{p-norm-table} compares different models trained with $p$-norm corruptions on the CIFAR-100 dataset with respect to the metrics described above. The table shows that all corruption-trained models produce improved $mCE$ and $mCE_{L_p}$ values at the expense of clean accuracy. Increasing the intensity of training corruptions mostly increases this effect. $L_{0}$ training is an exception where this effect is inconsistent. We find that there is a much smaller positive effect of corruption training on $mCE_{xN}$. For corruption training outside of $L_{0}$, increasing the intensity of corruptions actually worsens $mCE_{xN}$. The combined corruption models C1, C2 and C3 achieve much higher improvements in robustness than the models trained on one corruption only, with a similar degradation of clean accuracy. In particular, they achieve significantly improved $mCE_{xN}$ robustness. Model C2 performs better than C1 overall. C3 achieves slightly lower robustness to noise ($mCE$ and $mCE_{L_p}$), but better clean accuracy. The standard model and the models trained on $L_{0}$ corruptions show significant $iCE$ values, while $iCE$ is close to zero for all other models.

Table \ref{delta-table} illustrates the effect of a selection of training time $p$-norm corruptions and combined corruptions. To generalise the results, we report the delta of all metrics with the standard model, averaged over all model architectures. The table shows that $L_{0}$ corruption training slightly increases robustness, with the least negative impact on clean accuracy on the CIFAR datasets at the same time. The $L_{0.5}$ corruption training is the most effective among the single corruptions to increase $mCE$ and $mCE_{xN}$. The $L_\infty$ corruption training is the least effective in terms of the ratio of robustness gained per clean accuracy lost. In fact, on the Tiny ImageNet dataset, no single $p$-norm corruption training significantly reduces either $mCE$ or $mCE_{xN}$. On all datasets, C1, C2 and C3 achieve the most significant robustness improvements. C3 stands out particularly on Tiny ImageNet, where it is the only model that improves clean accuracy while effectively reducing $mCE$ or $mCE_{xN}$.

\begin{table}
\begin{center}
\setlength\tabcolsep{4.5pt}
{\caption{The effect of $p$-norm corruption training types relative to standard training, averaged across all model architectures and averaged for the two CIFAR datasets. Largest average improvement on the dataset for every metric is marked bold.}\label{delta-table}}
\resizebox{\linewidth}{!}{%
\begin{tabular}{lcccc}
\\[-10pt]
\hline
\rule{0pt}{12pt}
Model&$\Delta \overline{E_{clean}}$&$\Delta \overline{mCE}$&$\Delta \overline{mCE_{xN}}$&$\Delta \overline{mCE_{L_p}}$\\[3pt]
\hline
\rule{0pt}{13pt}
\textbf{CIFAR}&\\
\\[-9pt]
$L_{0}(0.01)$&\textbf{+0.3}&-4.3&-0.72&-8.96\\
$L_{0.5}$(7.5e+4)&+1.14&-6.15&-1.73&-10.62\\
$L_{2}(1)$&+1.12&-3.69&-1.18&-11.89\\
$L_\infty(0.02)$&+1.51&-2.74&-0.33&-10.9\\[7pt]
C1&+2.51&\textbf{-13.04}&\textbf{-4.95}&-27.42\\
C2&+1.79&-12.95&-4.78&\textbf{-27.86}\\
C3&+1.06&-10.8&-3.45&-25.82\\
\\[-9pt]
\hline
\rule{0pt}{13pt}
\textbf{TIN}&\\
\\[-9pt]
$L_{0}(0.02)$&+0.18&-0.8&+0.05&-4.06\\
$L_{0.5}$(1.2e+6)&-0.11&-0.22&+0.52&-4.39\\
$L_{2}(2)$&+0.02&+0.27&+0.82&-2.96\\
$L_\infty(0.04)$&+0.77&+0.95&+2.22&-4.47\\[7pt]
C1&+1.46&-2.57&-0.46&-14.67\\
C2&+0.41&\textbf{-2.81}&-0.74&\textbf{-15.52}\\
C3&\textbf{-0.14}&-2.69&\textbf{-0.79}&-12.13\\
\\[-9pt]
\hline
\end{tabular}}
\end{center}
\vspace{3pt}
\end{table}

\begin{figure}[t]
\centerline{\includegraphics[width=75mm, trim={5mm 0mm 14.5mm 11.5mm},clip]{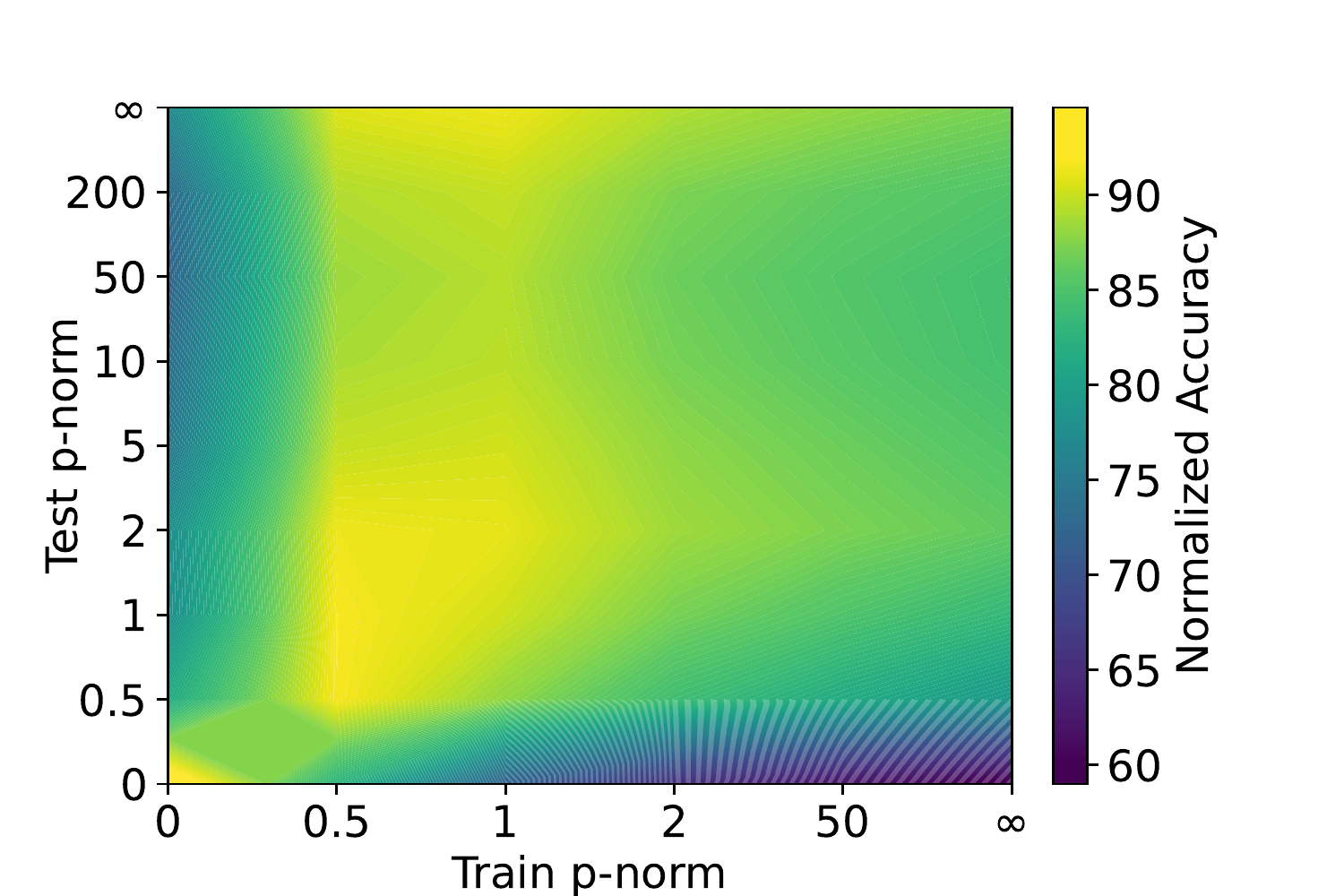}}
\caption{Normalized accuracy when training and testing on different $p$-norm corruptions. For each test corruption (see $mCE_{L_p}$ in Table \ref{corr-sets}), the accuracies of all models trained on $p$-norm corruptions (without additional data augmentation strategies) are first normalized so that the best model achieves 100\% accuracy. Then the average accuracy is calculated across all model architectures and datasets as well as across all $\epsilon$-values of the same $p$-norm for training and testing. This visualizes how, on average, training on one $p$-norm leads to robustness against all $p$-norm corruptions. The prior normalization makes the trained models comparable.} \label{2d-norm-gen}
\vspace{3pt}
\end{figure}

Figure \ref{2d-norm-gen} illustrates how the robustness obtained from training on different single $p$-norm corruptions transfers to robustness against other $p$-norm corruptions. It shows that $L_0$ corruptions are a special case. Training on $L_0$ corruptions gives the model very high robustness against $L_0$ corruptions, but less robustness against all other $p$-norm corruptions compared with other training strategies. At the same time, no $p$-norm corruption training except $L_0$ effectively achieves robustness against $L_0$ corruptions. Except for the $L_0$-norm, robustness from training on all other $p$-norm corruptions transfers across the $p$-norms. However, corruptions of lower $p$-norms such as $L_{0.5}$ and $L_1$ are more suitable for training than corruptions of higher norms such as $L_{50}$ and $L_\infty$. The former even achieves higher robustness against $L_\infty$-norm corruptions than $L_\infty$-norm training itself.

Table \ref{augmentation-table} illustrates the effectiveness of the data augmentation strategies RA, MU, AM and TA. RA and TA were not evaluated for corruption robustness in their original publications. The results show that all methods improve both clean accuracy and robustness. TA achieves the best $E_{clean}$ values. It leads to high robustness values on the CIFAR datasets and especially on the WRN model. AM is the most effective method overall in terms of robustness.
Our results show that MU, AM and TA can be improved by adding combined $p$-norm corruptions. For the CIFAR datasets, additional training with combined $p$-norm corruptions only leads to increases in $mCE$ and $mCE_{L_p}$, but not $mCE_{xN}$. On Tiny Imagenet, such additional corruptions can, in some combinations, increase $mCE_{xN}$ in addition to MU, AM and TA, without any loss of clean accuracy. 

All data augmentation strategies still achieve a significant $iCE$ value when not trained with additional corruptions. The $iCE$ value is much lower for Tiny Imagenet than for CIFAR.

\begin{table*}[t]
\begin{center}
\setlength\tabcolsep{1.5pt}
{\caption{The performance of state-of-the-art data augmentation techniques with additional $p$-norm corruption combinations. For WRN and RNX, we show those data augmentations that are most effective for at least one metric. The full results are available on Github.
}\label{augmentation-table}}
\resizebox{\textwidth}{!}{%
\begin{tabular}{l@{\hspace{9px}}ccccc@{\hspace{14px}}ccccc@{\hspace{13px}}ccccc}
\\[-10pt]
\hline
\rule{0pt}{10pt}
Model&$E_{clean}$&$mCE$&$mCE_{xN}$&$mCE_{L_p}$&$iCE$&$E_{clean}$&$mCE$&$mCE_{xN}$&$mCE_{L_p}$&$iCE$&$E_{clean}$&$mCE$&$mCE_{xN}$&$mCE_{L_p}$&$iCE$\\[3pt]
\rule{0pt}{10pt}
&\multicolumn{5}{c}{CIFAR-10}&\multicolumn{5}{c}{CIFAR-100}&\multicolumn{5}{c}{TinyImageNet}\\[2pt]
\cline{2-16}
\rule{0pt}{13pt}
\textbf{DN}&\\
\\[-9pt]
Standard&5.37 & 25.38 & 20.72 & 28.16 & 17.2\%   & 23.21 & 51.33 & 46.19 & 53.70 & 12.4\% & 37.38 & 76.46 & 75.28 & 53.82 & 1.4\% \\[5pt]
RA&4.59 & 18.25 & 14.91 & 21.76 & 19.0\% & 21.81 & 43.87 & 39.19 & 49.66 & 9.6\%  & 35.84 & 74.53 & 73.81 & 52.49 & 1.4\% \\[5pt]
MU&4.64 & 23.39 & 17.65 & 29.17 & 14.7\% & 22.16 & 48.96 & 42.67 & 53.22 & 7.7\%  & 35.13 & 72.81 & 71.49 & 50.43 & 0.6\% \\
MU+C1&5.68 & 13.89 & 15.44 & 7.03  & 2.1\%  & 23.92 & 36.96 & 39.39 & 26.21 & 0.5\%  & 36.75 & 71.40 & 72.10 & 38.86 & 0.2\% \\
MU+C2&5.36 & 13.77 & 15.28 & 6.98  & 3.5\%  & 23.40 & 36.88 & 39.31 & 25.95 & 0.6\%  & 35.87 & 70.79 & 71.35 & 38.05 & 0.5\% \\
MU+C3&4.92 & 14.65 & 15.35 & 8.42  & 1.9\%  & 22.84 & 38.22 & 39.50 & 28.17 & 0.2\%  & 35.27 & 71.20 & 72.12 & 42.47 & 0.4\% \\[5pt]
AM&4.88 & 13.14 & 11.69 & 13.21 & 4.6\%  & 23.27 & 37.90 & 35.85 & 37.11 & 3.6\%  & 36.80 & 67.03 & 66.42 & 49.13 & 1.9\% \\
AM+C1&5.94 & 11.32 & 12.15 & 7.14  & 1.1\%  & 25.17 & 34.86 & 36.37 & 27.56 & 1.0\%  & 38.15 & 65.19 & 65.65 & 40.13 & 0.2\% \\
AM+C2&5.30 & \textbf{10.58} & 11.34 & 6.75  & 2.1\%  & 24.98 & \textbf{34.50} & 35.96 & 27.40 & 1.2\%  & 38.10 & 65.02 & 65.30 & 40.17 & 0.2\% \\
AM+C3&5.08 & 11.34 & 11.48 & 8.15  & 3.6\%  & 24.30 & 35.32 & 35.81 & 29.24 & 1.0\%  & 37.08 & \textbf{64.12} & \textbf{65.12} & 43.20 & 0.5\% \\[5pt]
TA&\textbf{4.43} & 14.25 & \textbf{11.12} & 17.08 & 8.4\% & \textbf{20.04} & 37.85 & \textbf{33.25} & 42.71 & 9.0\% & 34.46 & 72.61 & 72.39 & 72.36 & 0.7\% \\
TA+C1&5.19 & 11.94 & 13.07 & \textbf{6.63}  & 3.3\%  & 22.74 & 35.27 & 37.34 & 25.52 & 1.2\%  & \textbf{33.99} & 68.97 & 69.58 & \textbf{36.27} & 0.2\% \\
TA+C2&5.01 & 12.88 & 14.21 & 6.68  & 2.3\%  & 22.43 & 35.32 & 37.40 & \textbf{25.34} & 1.4\%  & 34.40 & 69.91 & 70.43 & 36.90 & 0.5\% \\
TA+C3&4.88 & 13.36 & 14.16 & 7.63  & 1.8\%  & 21.92 & 36.36 & 37.82 & 26.61 & 0.5\%  & \textbf{34.01} & 71.48 & 73.46 & 40.53 & 0.4\% \\
\\[-8pt]
\hline
\rule{0pt}{13pt}
\textbf{WRN}&\\
\\[-9pt]
RA & 4.56      & 21.80     & 18.45     & 24.02      & 12.1\%    & 23.25     & 48.65     & 44.77     & 50.96     & 7.8\%    & \textbf{36.89}     & 75.2     & 73.99     & 54.05     & 3.0\% \\
AM+C2 & 5.53      & \textbf{13.39}     & 14.72     & \textbf{7.13}      & 2.1\%    & 25.40     & \textbf{37.08}     & 39.03     & \textbf{28.01}     & 0.6\%    & 38.48     & \textbf{67.34}     & \textbf{67.46}     & \textbf{41.22}     & 0.5\% \\
TA&\textbf{4.13}      & 15.45     & \textbf{11.87}     & 18.68     & 7.0\%    & \textbf{21.70}     & 41.87     & \textbf{37.81}     & 44.32     & 8.2\%    & 38.25     & 78.18     & 77.92     & 77.88     & 1.6\%  \\
\hline
\rule{0pt}{13pt}
\textbf{RNX}&\\
\\[-9pt]
AM & 4.35 & 14.06 & \textbf{12.13} & 14.94 & 9.2\%  & 21.45 & 38.10 & 35.31 & 39.39 & 4.5\%  & 34.97     & 67.79     & 66.09     & 47.85     & 1.7\% \\
AM+C2& 5.87 & \textbf{11.83} & 12.87 & 7.25  & 2.0\%   & 24.22 & \textbf{33.95} & 35.60 & 26.59 & 0.8\%  & 35.03     & \textbf{65.01}     & \textbf{65.52}     & 37.51     & 0.7\% \\[5pt]
TA & \textbf{4.14} & 15.56 & 12.55 & 18.62 & 10.5\%  & \textbf{19.55} & 37.80 & \textbf{34.28} & 40.44 & 6.6\%  & 31.70     & 75.27     & 74.98     & 74.95     & 2.0\%\\
TA+C2& 5.27 & 14.09 & 15.86 & \textbf{6.72}  & 0.7\%   & 21.28 & 35.09 & 37.52 & \textbf{24.48} & 1.5\%  & \textbf{31.49}     & 71.13     & 71.99     & \textbf{34.20}     & 0.9\% \\
\\[-9pt]
\hline
\end{tabular}}
\end{center}
\vspace{8pt}
\end{table*}

\begin{figure}[t]
\centerline{\includegraphics[width=75mm, trim={42mm 11mm 32mm 11mm},clip]{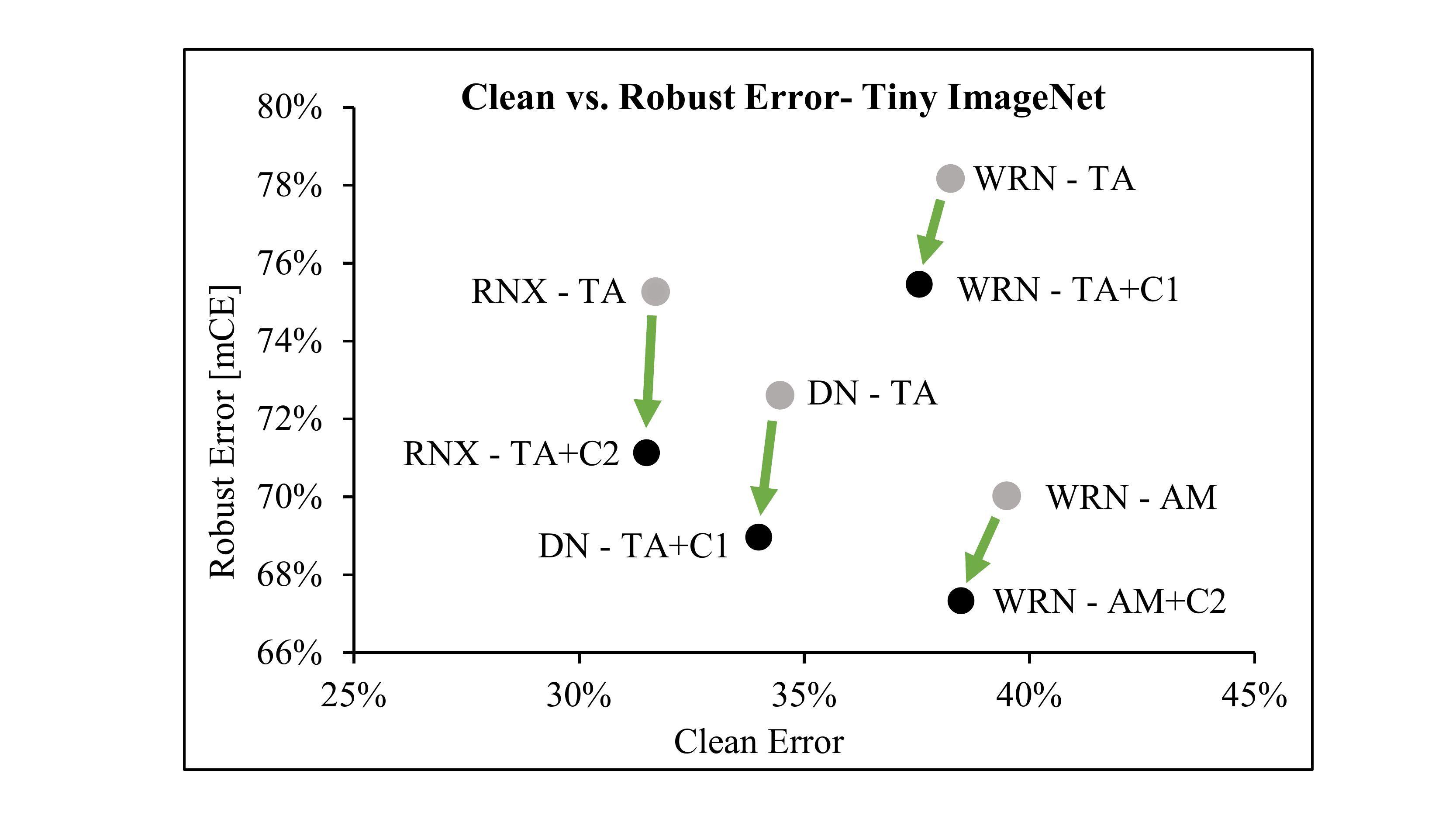}}
\caption{Clean Error vs. $mCE$ plot for selected models on Tiny Imagenet. The green arrows indicate an improvement of both metrics when the model is trained on $p$-norm corruption combinations.} \label{acc-rob}
\vspace{4pt}
\end{figure}

Figure \ref{acc-rob} visualizes $E_{clean}$ vs. $mCE$ for selected model pairs. One of the pairs is additionally trained with C1 or C2. Models towards the lower left corner are both more accurate and more robust. The arrows show how C1 and C2 training can mitigate the trade-off between accuracy and robustness in these cases.

Figure \ref{learningcurve} in the appendix shows how training data augmentation has an implicit regularizing effect on the training process, flattening the learning curve. This effect is evident for the C1 combination, although much less significant than for TA. Strong random training data augmentation can allow for a longer training process \cite{Vryniotis2021}.

\section{\uppercase{Discussion}}

\textbf{Vulnerability to Imperceptible Corruptions} \quad For the CIFAR datasets in particular, we found $iCE$ values of well above $10\%$. Models trained with state-of-the-art data augmentation methods are still equally vulnerable to imperceptible random noise. Training with arbitrary noise outside $L_0$ corruptions solves the problem. We recommend that the set of imperceptible corruptions be further developed as a minimum requirement for the robustness of vision models, and that metrics such as $iCE$ be evaluated for this purpose.

\textbf{Robustness transfer to Real-World Corruptions} \quad From our experiments we derive insights into improving real-world robustness by training with $p$-norm corruptions: 
\begin{itemize}
\item Training with single $p$-norm corruptions generally leads to $mCE$ improvements only on CIFAR, not on Tiny ImageNet. The improvements are mainly attributable to the noise types within $mCE$. There is little to negative transfer of robustness against corruption types outside the pixel-wise noise, represented by the $mCE_{xN}$ metric.
\item Training with combinations of $p$-norm corruptions leads to robustness improvements for the majority of corruption types outside of pixel-wise noise\footnote{Visit Github for all individual results}. Thus, $mCE_{xN}$ improves even on Tiny ImageNet. 
\item RA, MU, AM and TA show significant robustness improvements against all real-world corruptions. 
\item Adding $p$-norm corruption combinations to RA, MU, AM and TA gives mixed results. While $mCE$ often improves, $mCE_{xN}$ does so less often. For Tiny ImageNet, C2 improves $mCE_{xN}$ quite effectively on top of all other data augmentation strategies. For CIFAR, $mCE_{xN}$ could not be improved significantly beyond AM and TA.
\end{itemize}

\textbf{Robustness transfer across $p$-norms} \quad Given the limited transferability of robustness between different types of corruptions known from the literature and the differences in volume for different $p$-norm balls (Table \ref{volumefactors}), we expected models trained on one $p$-norm to achieve high robustness against that $p$-norm in particular. However, we found that training on $p$-norm corruptions outside of $L_0$ leads to good robustness against all $p$-norm corruptions outside of $L_0$. We find that training on $L_\infty$-norm corruptions performs worse overall compared to other $p$-norms. The $L_0$ corruption seems to be a special case that generalizes mainly to itself. We investigate this behaviour by examining whether $p-norm$ corruptions overlap in the input space (see Figure \ref{comparevolumes} in the appendix).

We find that the random $L_0$ corruptions are a special case in that they almost never share the same input space with an $L_2$-norm ball, and vice versa. In a weaker form, this is also true for other norms with dissimilar $p$ like $L_\infty$-norm and $L_2$. However, when comparing corruptions from more similar $p$-norms like $L_1$-norm and $L_2$-norm, we find that for most $\epsilon$ values the corruptions share the same input space. The observations from Figure \ref{comparevolumes} explain, from a coverage perspective, why corruption robustness seems to transfer between different $p$-norms outside of $L_0$. The $L_\infty$-norm is a corner case with the highest $p$-norm and, similarly to $L_{50}$, appears somewhat distinct from $L_2$ in Figure \ref{comparevolumes}. Interestingly, our experiments still show that robustness against high-norm corruptions is most effectively achieved by training on other $p$-norm corruptions.

From the Figures \ref{2d-norm-gen} and \ref{comparevolumes} we conclude that it makes little difference to train and test on a large variety of random $p$-norm corruptions with $p>0$. This is especially true for arbitrarily chosen $\epsilon$-values like the set of corruptions for the $mCE_{L_p}$-metric. These corruptions will mostly be sampled in the same region of the input space and, therefore, be largely redundant.

\textbf{Promising Data Augmentation Strategies} \quad From our investigations of robustness transfer across $p$-norms we conclude that the C2 strategy is generally more useful than C1 because it contains only one $p$-norm corruption with $0<p<\infty$, whereas additional such corruptions would be redundant. We find evidence for this assumption in our experimental results, where on average C2 achieves similar robustness gains at a lower cost of clean accuracy. In future work, we would like to investigate whether the C2 combination can be further improved. First, the ineffective $L_\infty$ corruption could be reduced in impact. Second, $L_2$ could be replaced by the more effective $L_{0.5}$ or $L_1$-norm corruptions or by simple Gaussian noise.

All data augmentation methods MU, RA, AM and TA improve both corruption robustness and clean accuracy and are therefore highly recommended for all models and datasets. Therefore, the most relevant question is whether $p$-norm corruption training can be combined with these methods. Our experiments show a mixed picture: Robustness against noise can be improved, but robustness against other corruptions less so. The results vary across datasets, model architectures, and base data augmentation methods, as well as across the different severities of $p$-norm corruptions (C1 vs C3). Future progress on this topic requires precise calibration of the $p$-norm corruption combinations and possibly multiple runs of the same experiment to account for the inherent randomness of the process and to obtain representative results. In future work, we aim to improve the $p$-norm corruption combinations and apply them for more advanced training techniques as in \cite{Lim2021,Erichson2022}. We believe that this study provides arguments in principle that corruption combinations are more effective than single noise injections. 

\section{\uppercase{Conclusion}}
Robustness training and evaluation with random $p$-norm corruptions has been little studied in the literature. We trained and tested three classification models with random $p$-norm corruptions on three image datasets. We discussed how robustness transfers across $p$-norms from an empirical and test coverage perspective. The results show that training data augmentation with $L_0$-norm corruptions is a specific corner case. Among all other $p$-norm corruptions, lower $p$ are more effective for training models. Combinations of $p$-norm corruptions are most effective, which can achieve robustness against corruptions other than pixel-wise noise. Depending on the setup, $p$-norm corruption combinations can improve robustness when applied in sequence with state-of-the-art data augmentation strategies. We investigated three different $p$-norm corruption combinations and, based on our findings, made suggestions for further improving robustness. Our experiments show that several models, including those trained with state-of-the-art data augmentation techniques, are negatively affected by quasi-imperceptible random corruptions. Therefore, we emphasized the need to evaluate the robustness against such imperceptible corruptions and proposed an appropriate error metric for this purpose. In the future, we plan to further improve robustness by more advanced training data augmentation with corruption combinations.

\bibliographystyle{apalike}
{\small
\bibliography{example}}

\section*{\uppercase{Appendix}}

\subsection*{Volume of $p$-norm balls}

\begin{table}[htb]
\begin{center}
{\caption{Volume factors between $L_\infty$-norm ball and $L_2$-norm ball as well as $L_2$-norm ball and $L_1$-norm ball of the same $\epsilon$ in $d$-dimensional space}\label{volumefactors}}
\begin{tabular}{lcc}
\\[-9pt]
\hline
\rule{0pt}{12pt}
$d$&$p=[\infty, 2]$&$p=[2, 1]$\\
\\[-9pt]
\hline
\\[-7pt]
~3\quad&\quad1.9&3.1\\
~5\quad&\quad19.7&6.1\\
~10\quad&\quad9037&401.5\\
~20\quad&\quad$6*10^{10}$&$4*10^7$\\
\\[-8pt]
\hline
\\[-15pt]
\end{tabular}
\end{center}
\end{table}

\subsection*{Sampling Algorithm}

We use the following sampling algorithm for norms $0 < p < \infty$, which returns a corrupted image $I_c$ with maximum distance $\epsilon$ for a given clean image $I$ of any dimension $d$:

\begin{enumerate}
    \item Generate $d$ independent random scalars $x_i=(G(1/p,1))^{1/p}$, where $G(1/p,1)$ is a Gamma-distribution with shape parameter $1/p$ and scale parameter $1$.
    \item Generate $I$-shaped vector $x$ with components $x_i*s_i$, where $s_i$ are random signs.
    \item Generate scalar $r = w^{1/d}$ with $w$ being a random scalar drawn from a uniform distribution of interval $[0,1]$.
    \item Generate $n = (\sum\nolimits_{i=1}^{d}|x_i|^{ p })^{ 1/p }$ to norm the ball.
    \item Return $I_c = I+(\epsilon*r*x/n)$ 
\end{enumerate}

The factors $r$ and $w$ allow to adjust the density of the data points radially within the norm ball. We generally use a uniform distribution for $w$ except when we sample imperceptible corruptions (Figure \ref{impercept}).  
\\
\subsection*{Volume overlap of $p$-norm balls}

\begin{figure*}[btp]
\centerline{\includegraphics[width=\linewidth, trim={39mm 13mm 45mm 24mm},clip]{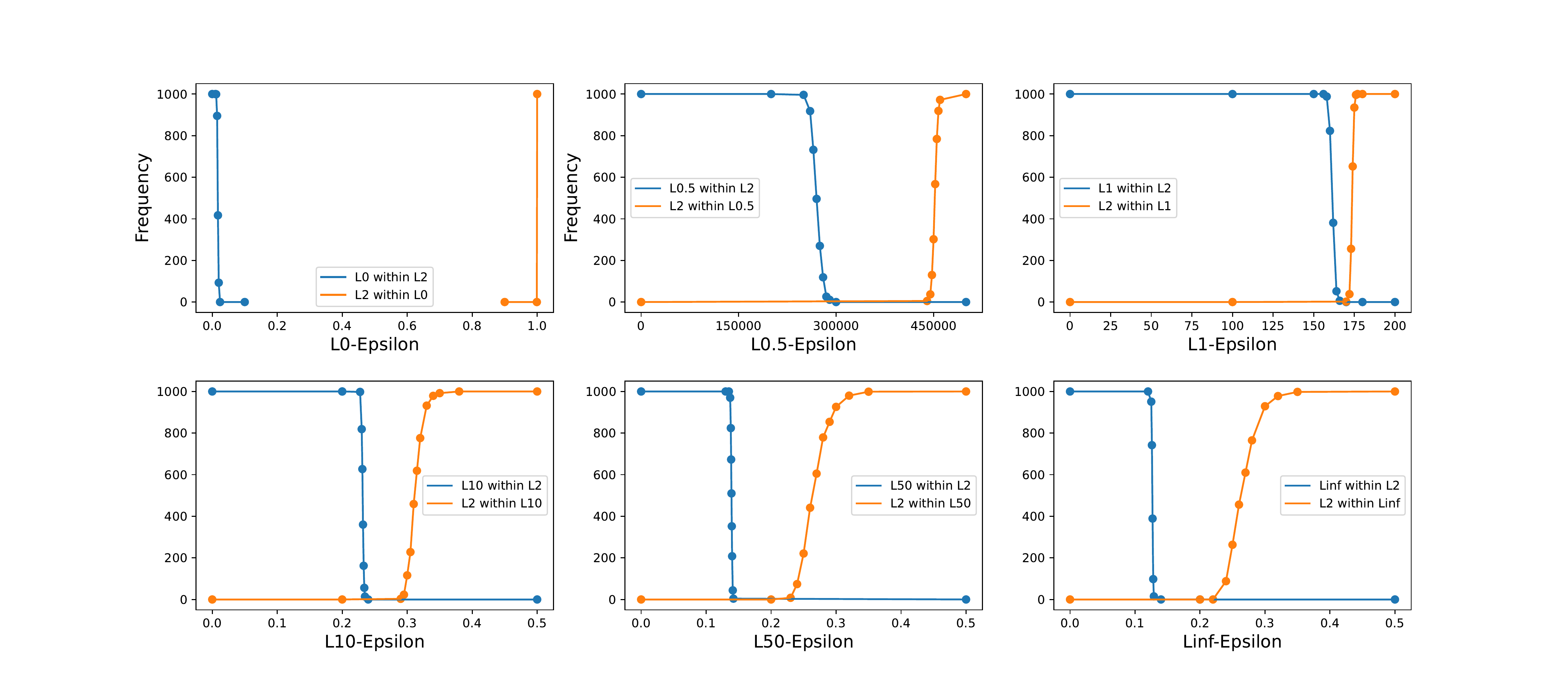}}
\caption{The frequency of 1000 samples drawn from inside a first CIFAR-10-dimensional $p$-norm ball also being part of a second $L_2$-norm ball of $\epsilon=4$ (blue plot), as well as the frequency of 1000 samples drawn from inside the second norm ball also being part of the first norm ball (orange plot).} 
\label{comparevolumes}
\vspace{5pt}
\end{figure*}

In Figure \ref{comparevolumes} we estimate the overlap of the volumes of two norm balls with different $p$ and with a dimensionality equal to CIFAR-10 (3072). We estimate their volumes by uniformly drawing 1000 samples from inside the norm ball. The figure shows 6 sub-plots for different first norm balls, where their respective $\epsilon$ being varied along the x-axis. The blue plots indicate how many samples from this first $p$-norm ball are also part of a second $L_2$-norm ball with $\epsilon=4$. Similarly, the orange plots show how many samples from the second $L_2$-norm ball are also part of the first norm ball.

There is a large interval of $L_0-\epsilon$-values, where the samples do not overlap. In a weaker form, this is also true for other norms far away from $p=2$, like the $L_\infty$-norm and the $L_{0.5}$-norm. This means that when comparing $L_2$-norms with e.g. $L_0$-norms, there is a large range of $\epsilon$-values, where the two norm balls cover predominantly different regions of the input space. However, the more similar the $p$-norms compared get, like $L_1$ with $L_2$, a different result can be observed. One of the norm balls predominantly overlaps the other in volume for the widest part of $\epsilon$-values. In such a case, the covered input space is mostly redundant.

\subsection*{Learning Curve Effect}

\begin{figure}
\centerline{\includegraphics[width=75mm, trim={6.5mm 11mm 15mm 20mm},clip]{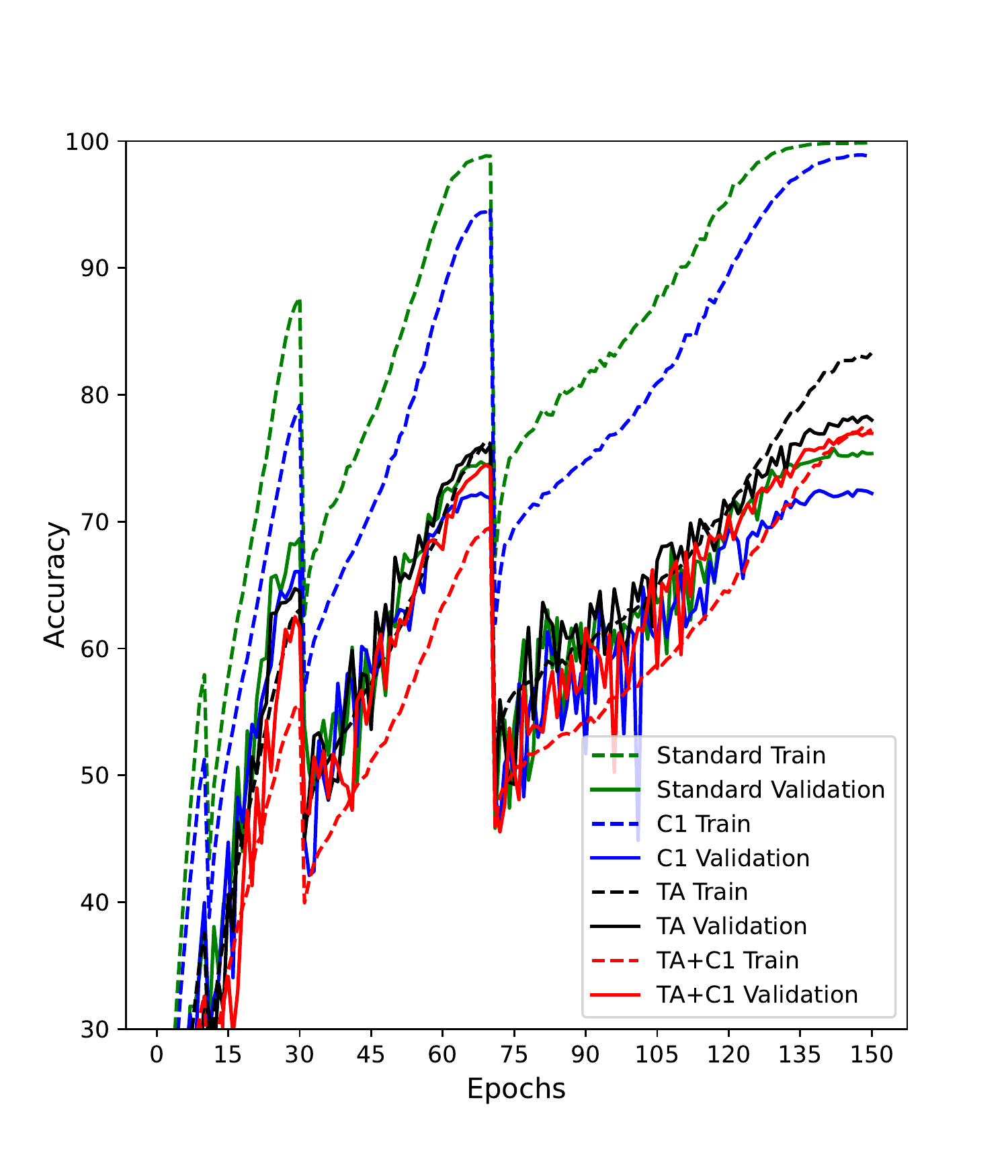}}
\caption{Training and validation learning curves of various models show the slight regularizing effect of $p$-norm corruptions in the C1 combination on the training process.} \label{learningcurve}
\vspace{10pt}
\end{figure}

When added to a standard or TA training procedure, the training curve of the C1 model is more flat (Figure \ref{learningcurve}. TA itself shows a very strong  flattening effect on the learning curve and requires visibly more epochs to converge.

\end{document}